\documentclass[11pt,a4paper]{article}
\usepackage{authblk}

\usepackage[hyperref]{emnlp-ijcnlp-2019}
\usepackage{times}
\usepackage{latexsym}

\usepackage{url}
\usepackage{bm}
\definecolor{Gray}{gray}{0.8}
\usepackage{graphicx}
\usepackage{amsmath}
\usepackage{booktabs}
\usepackage{algorithm}
\usepackage{algorithmic}
\usepackage{makecell}
\usepackage{color, colortbl}
\usepackage{pbox}
\usepackage{multirow}

\newcommand{\E}{\mathrm{E}}

\newcommand{\rlbleu}{GE$_{\textnormal{BLEU}}$}
\newcommand{\rlbleuqa}{GE$_{\textnormal{BLEU+QSS+ANSS}}$}
\newcommand{\rlgleuqa}{GE$_{\textnormal{GLEU+QSS+ANSS}}$}
\newcommand{\rlrougeqa}{GE$_{\textnormal{ROUGE+QSS+ANSS}}$}
\newcommand{\rldasqa}{GE$_{\textnormal{DAS+QSS+ANSS}}$}

\newcommand{\rlgleu}{GE$_{\textnormal{GLEU}}$}
\newcommand{\rldas}{GE$_{\textnormal{DAS}}$}
\newcommand{\rlrouge}{GE$_{\textnormal{ROUGE}}$}

\newcommand{\sumbleu}{SUM$_{\textnormal{BLEU}}$}
\newcommand{\sumrouge}{SUM$_{\textnormal{ROUGE}}$}

\aclfinalcopy 

\title{Putting the Horse Before the Cart:
\\  A Generator-Evaluator Framework for Question Generation from Text}
\author[1,2,3]{Vishwajeet Kumar}
\author[2]{Ganesh Ramakrishnan}
\author[3]{Yuan-Fang Li}
\affil[1]{IITB-Monash Research Academy, Mumbai, India}
\affil[2]{IIT Bombay, Mumbai, India}
\affil[3]{Monash University, Melbourne, Australia}
\date{}
\begin{document}
\maketitle


\begin{abstract}
Automatic question generation (QG) is a useful yet challenging task in NLP. Recent neural network-based approaches represent the state-of-the-art in this task. In this work, we attempt to strengthen them significantly by adopting a holistic and novel generator-evaluator framework that directly optimizes objectives that reward semantics and structure. 
The {\it generator} is a sequence-to-sequence model that incorporates the {\it structure} and {\it semantics} of the question being generated. The generator predicts an answer in the passage that the question can pivot on. Employing the copy and coverage mechanisms, it also acknowledges other contextually important (and possibly rare) keywords in the passage that the question needs to conform to, while not redundantly repeating words.
The {\it evaluator} model evaluates and assigns a reward to each predicted question based on its conformity to the {\it structure} of ground-truth questions. We propose two novel QG-specific reward functions for text conformity and answer conformity of the generated question. The evaluator also employs structure-sensitive rewards based on evaluation measures such as BLEU, GLEU, and ROUGE-L, which are suitable for QG. In contrast, most of the previous works only optimize the cross-entropy loss, which can induce inconsistencies between training (objective) and testing (evaluation) measures.
Our evaluation shows that our approach significantly outperforms state-of-the-art systems on the widely-used SQuAD benchmark as per both automatic and human evaluation.
\end{abstract}


\section{Introduction}
\label{Intro}
Automatic question generation (QG) is a very important yet challenging problem in NLP. It is defined as the task of generating syntactically correct, semantically sound and relevant questions from various input formats such as text, a structured database or a knowledge base \cite{mannem2010question}. 
More recently, neural network based techniques such as sequence-to-sequence (Seq2Seq) learning have achieved remarkable success on various NLP tasks, including question generation. A recent deep learning approach to question generation  \cite{serban2016generating} investigates a simpler task of generating questions only from a triplet of subject, relation and object.  Learning to ask (referred to as L2A hereinafter)~\cite{du2017learning} proposes a Seq2Seq model with attention for question generation from text. \cite{song2018leveraging} (in an approach referred to as NQG$_{LC}$ hereafter) encoded ground-truth answers and employed bi-directional LSTMs in a Seq2Seq setting. In addition, they use the {\em copy} mechanism~\cite{see2017get} and context matching to capture interactions between the given ground-truth answer and its context within the passage.

In the context of QG from paragraphs, \cite{zhao2018paragraph} proposed maxout pointer network to keep track of word {\em coverage}. 
Our previous work ~\cite{kumarpakdd2018} (referred to as AutoQG hereinafter) generates {\em candidate answers} from text using Pointer Networks~\cite{vinyals2015pointer} and {\em encodes the answer} in the question decoder for improved performance.

We first present a framework in which a {\em generator} mechanism (the horse) that is employed  for generating a question-answer pair invokes or pulls the {\em evaluator} mechanism (the cart) that is employed for evaluating the generated pair. Our clearly delineated {\em generator}-{\em evaluator} framework lets us (a) easily incorporate several best practices 
from the above referred previous models  in the {\em generator} while (b) also letting us employ in the {\em evaluator}, other complex non-decomposable rewards that are consistent with performance measures (such as BLEU and ROUGE) on test data. We also propose some novel reward functions that {\em evaluate} the syntax of the question and semantics of the question-answer pair in its entirety.
More specifically, since the generated question is in anticipation of some specific answer, we find it most natural to incorporate candidate answer generation (using  Pointer Networks) alongside QG right in our generator module, so that the evaluator can optionally take into cognizance the conformity of the generated answer to the ground-truth answer, along with text conformity. Likewise, we also incorporate copy and coverage mechanisms for QG into the generator module so that they can be specifically trained by leveraging a suite of holistically designed and structure-sensitive  reward functions in the evaluator module.
  
\begin{table*}[htb]
\centering
\small
\scalebox{0.8}{
\begin{tabular}{|l|l|l|} 
\hline
\multicolumn{3}{|c|}{\parbox[t][30pt][t]{.96\linewidth}{\textbf{Text:} 
``new york city traces its roots to its 1624 founding as a trading post by colonists of the dutch republic and was named new amsterdam in 1626 .''}}\\
\hline
\textbf{Row} & \textbf{Model} &  \textbf{Question generated} \\
\hline
1 & Seq2Seq model optimized on vanilla (cross entropy) loss without answer prediction & in what 1624 did new york city traces its roots ?\\
\hline
2 & Seq2Seq model optimized on vanilla (cross entropy) loss with answer prediction & what year was new york named ?\\
\hline
3 & Copy aware Seq2Seq model & what year was new new  amsterdam  named ? \\ \hline
4 & Coverage and copy aware Seq2Seq model  & in what year was new amsterdam named ? \\ \hline
5 & Seq2Seq model optimized on BLEU (using RL) &  what year was new york founded ? \\ \hline
\end{tabular}}
\caption{Sample text and questions generated using variants of our model.}\label{tab:sample}
\end{table*}

\begin{table*}[htb]
\centering 
\scalebox{0.8}{
\begin{tabular}{|l|l|l|} 
\hline
\multicolumn{3}{|l|}{\parbox[t][48pt][t]{\linewidth}{\textbf{Text:} 
``even with the five largest cities in sichuan suffering only minor damage from the quake , some estimates of the economic loss run higher than us \$ 75 billion , making the earthquake one of the costliest natural disasters in chinese history .''}}\\
\multicolumn{3}{|l|}{\textbf{Expected answer:} five}\\
\hline
\textbf{Row} & \textbf{Model} &  \textbf{Question generated} \\
\hline
1 &  \rlbleu & how much did it making for the earthquake of the economic ? \\ \hline
2 & \rlbleuqa &  how many largest cities in sichuan experience only minor damage from the quake ? \\ \hline
3 &  \rldas &  how many cities were in sichuan ? \\ \hline
4 & \rldasqa & how many largest cities in sichuan suffering only minor damage from the quake ? \\
\hline
4 & \rlrouge &  how much did the economic loss run in sichuan ? \\ \hline
5 & \rlrougeqa & what is the largest cities in sichuan ? \\
\hline
\end{tabular}}
\caption{Sample text and questions generated using different reward functions, with and without our new QG-specific rewards QSS+ANSS.}\label{tab:samplerl}
\end{table*}

\subsubsection*{The Generator}
In Table~\ref{tab:sample}, in rows 1 through 4, we illustrate through examples, the incremental benefits of introducing answer prediction and the copy and coverage mechanisms \cite{see2017get} in the generator. The evaluator associated with the corresponding three generator models employs the conventional and simplistic cross-entropy loss. The motivation for answer prediction in the generator module is obvious and will be further discussed in Section 2.1. In row 3 we illustrate the influence of our copy mechanism, where a rare phrase `new amsterdam' has been rightly picked up in association with the name of the city.

We however note that in row 3, the word `new' has been erroneously repeated twice, since an encoder-decoder based model could generate questions with meaningless repetitions.

We introduce a mechanism for discouraging such repetitions in our generator by quantitatively emphasizing the \emph{coverage} of sentence words while decoding. Row 4 shows the improved and relevant question generated by our model trained by incorporating both the copy and coverage mechanisms.

\subsubsection*{Evaluator}
In row 5 of Table~\ref{tab:sample}, we observe the high-quality question that is generated when the simplistic cross-entropy loss in the evaluator is replaced with the more complex and non-decomposable (across words) BLEU reward that accounts for proximity of `founded' to `new york'.

In Table~\ref{tab:samplerl}, we further illustrate the effect of employing other reward functions (described in Section 2.2) in the evaluator. As can be seen, the model that incorporates QG-specific reward functions (QSS and ANSS) generates a significantly better question when compared to the question generated without these rewards.

\noindent \textbf{Limitations of simple decomposable losses:}
A Seq2Seq model trained using a vanilla  cross-entropy loss function (decomposable over words in the question) generates the question \textit{``what year was new york named ?''} (row 1 in Table~\ref{tab:sample}), which is not addressed in the sentence. The passage talks only about the founding of the city  and its naming two years later. The inaccuracy of the question is possibly caused by the use of a loss that is agnostic to sequence information. In other words, given its decomposable nature, the cross-entropy loss on the ground-truth question or any of its (syntactically invalid) anagrams will be the same.
Moreover, use of the cross-entropy loss in the sequence prediction model could make the process brittle, since the model trained on a specific distribution over words is used on a test dataset with a possibly different distribution to predict the next word given the current predicted word. This creates exposure bias~\cite{ranzato2015sequence} during training, since the model is only exposed to the data distribution and not the model distribution. Thus, performance suffers due to inadequately evaluating the \emph{structure} of the generated question against the ground-truth question.

The standard metrics for evaluating the performance of question generation models such as BLEU~\cite{papineni2002bleu}, GLEU, and ROUGE-L~\cite{lin2004rouge} are based on degree of n-gram overlaps between a generated question and the ground-truth question. It would be desirable to be able to directly optimize these \emph{task-specific metrics}. 
However, these n-gram based metrics do not decompose over individual words and are therefore hard to optimize. 
We explicitly employ an evaluator that rewards each generated question based on its conformance to one (or more than one using decomposable attention) questions in the ground-truth set using these possibly non-decomposable reward functions. We find such learning to be a natural instance of reinforcement learning (RL)~\cite{sutton1998introduction} that allows us to use policy gradient to directly optimize task-specific rewards (such as BLEU, GLEU and ROUGE-L), which are otherwise non-differentiable and hard to optimize. In Table \ref{tab:samplerl} we illustrate questions generated using different reward functions. It can be observed that questions generated using combination of standard reward functions with reward functions specific to QG quality (QSS+ANSS) exhibit higher quality.

\paragraph{Contributions}
We summarize our main contributions as follows: 
\begin{itemize}
\item A comprehensive, end-to-end \textbf{generator-evaluator framework} naturally suited for automated question generation. Whereas earlier approaches employ some mechanism (the horse) for generating the question, intertwined with an evaluation mechanism (the cart), we show that these approaches can benefit from a much clearer separation of the generator of the question from its evaluator.
\item A {\em generator} founded on the  \textbf{semantics} and \textbf{structure} of the question  by (a) identifying target/pivotal answers (Pointer Network), (b) recognizing contextually important keywords in the answer (copy mechanism), and (c) 
avoiding redundancy (repeated words) in the question (coverage mechanism).
\item An {\em evaluator} that (a) directly optimizes for conformity to the \textbf{structure} of ground-truth sequences (BLEU, GLEU, etc.), and (b) matches against appropriate questions from a set of ground-truth questions (Decomposable Attention).
\item Novel reward functions that ensure that the generated question is relevant to the text and conforms to the  encoded answer.
\end{itemize}
When evaluated on the benchmark SQuAD dataset~\cite{rajpurkar-EtAl:2016:EMNLP2016}, our system considerably outperforms state-of-the-art question generation models~\cite{du2017learning,kumarpakdd2018,song2018leveraging} in automatic and human evaluation. 


\section{Our Approach}
\label{ourapp}
Our framework for question generation consists of a generator and an evaluator. From the reinforcement learning (RL) point of view, the generator is the \textit{agent} and the generation of the next word is an \textit{action}. The probability of decoding a word $P_\theta(word)$ gives a stochastic \textit{policy}. On every token that is output, an evaluator assigns a reward for the output sequence predicted so far using the current policy of the generator. Based on the reward assigned by the evaluator, the generator updates and improves its current policy. Let us denote the reward (\textit{return}) at time step $t$ by $r_t$. The cumulative reward, computed at the end of the generated sequence is represented by $R = \sum_{t=0}^T r_t$. 
The goal of our framework is to determine a generator (policy) that maximizes the expected return:
\begin{equation}
\label{rlloss}
Loss_{RL}(\theta) = - \E_{P_\theta(Y_{0:T}|\textbf{X})} \sum\limits_{t=0}^{T}r_t(Y_t;\textbf{X},Y_{0:t-1})
\end{equation}
where $X$ is the current input and $Y_{0:t-1}$ is the predicted sequence until time $t-1$. This supervised learning framework allows us to directly optimize task-specific evaluation metrics ($r_t$) such as BLEU. 

\begin{figure}
\centering
\includegraphics[width=\linewidth]{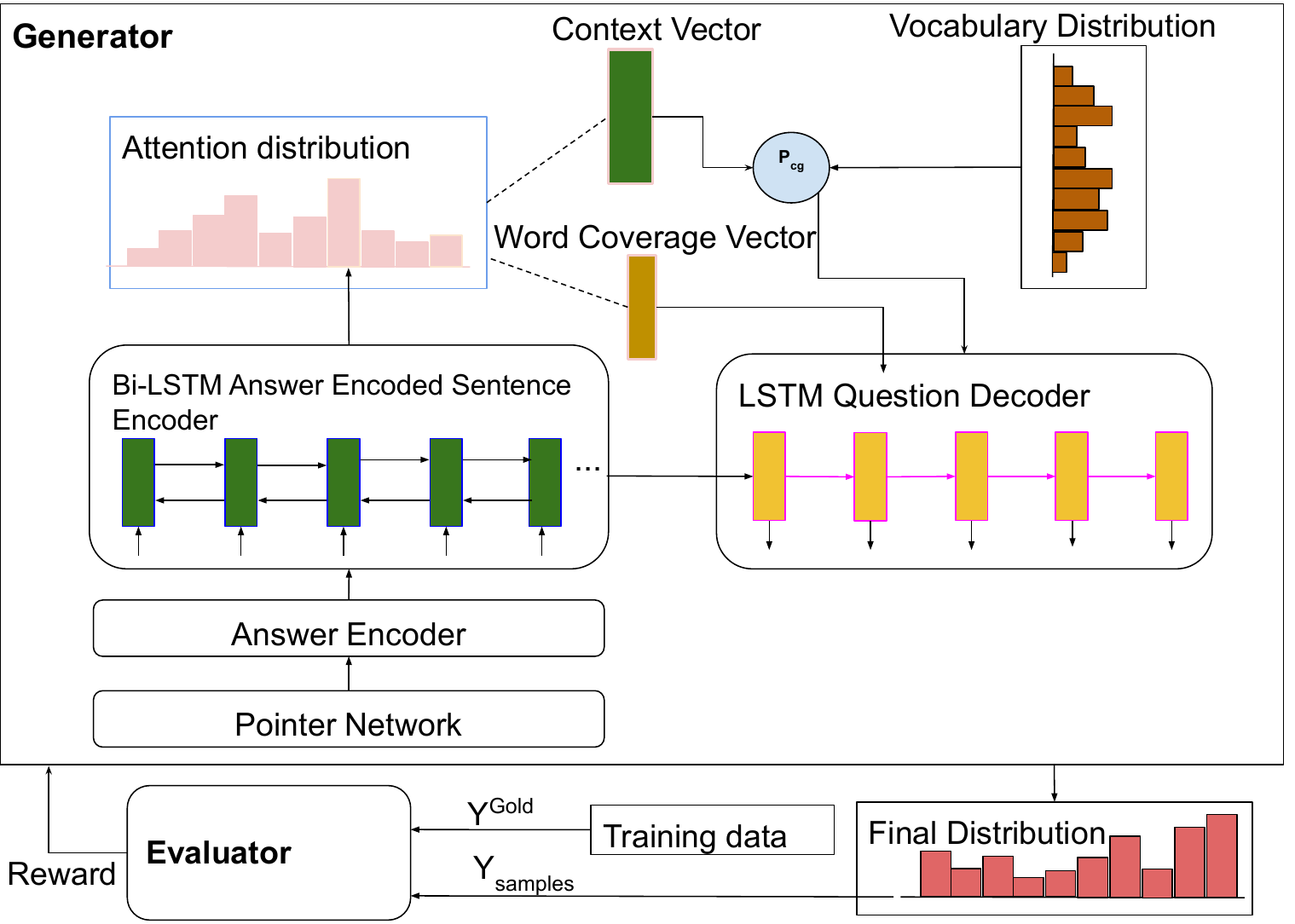}
\caption{Our generator-evaluator framework for question generation. $p_{cg}$ is the probability which determines whether to copy a word from source text or sample it from vocabulary distribution.}
\label{arch}
\end{figure}

 The generator is a sequence-to-sequence model, augmented with (i) an encoding for the potentially best pivotal answer, (ii) the copy mechanism~\cite{gu2016incorporating} to help generate contextually important words, and (iii) the coverage mechanism~\cite{DBLP:conf/acl/TuLLLL16} to discourage word repetitions. The evaluator provides rewards to fine-tune the generator. The reward function can be chosen to be a combination of one or more metrics. 
The high-level architecture of our question generation framework is presented in Figure~\ref{arch}.

\subsection{Generator}
\label{gen}
Similar to AutoQG~\cite{kumarpakdd2018}, we employ attention and boundary pointer network to identify pivotal answer spans in the input sentence. The generator then takes as input the sequence of words in the sentence, each augmented  with encoding of most probable pivotal answer, along with a set of linguistic features such as POS tag, NER tag, {\it etc.} At each step, the generator outputs a word with the highest probability, to eventually produce a word sequence. Additionally, as we will see, the generator employs copy and coverage mechanisms. 

\paragraph{Sentence Encoder:} Each word in the input text is fed sequentially into the encoder along with its linguistic features as well as with the encoded pivotal answer (identified by the boundary pointer network).
Our encoder is a two-layer bidirectional LSTM network, consisting of
$\overrightarrow{h_t} = \overrightarrow{LSTM_2}(x_t,\overrightarrow{h_{t-1}})$ and 
$\overleftarrow{h_t}  = \overleftarrow{LSTM_2}(x_t,\overleftarrow{h_{t-1}})$, 
which generates a sequence of hidden states.
Here $x_t$ is the given input word at time step $t$, and $\overrightarrow{h_t}$ and $\overleftarrow{h_t}$ are the hidden states at time step $t$ for the forward and backward passes respectively. 

\paragraph{Question Decoder:} Our question decoder is a single-layer LSTM network, initialized with the state $s= [\overrightarrow{h_t}; \overleftarrow{h_t}]$, which is concatenation of hidden state from forward and backward passes.

We also model the attention~\cite{bahdanau2014neural} distribution over words in the source text. 
We calculate the attention ($a^t_i$)  over the $i^{th}$ source word as $a^t_i = softmax(e^t_i)$, where 
\begin{align}
\label{att}
e^t_i &= v^ttanh(W_{eh}h_i+W_{sh}s_t+b_{att})
\end{align}
Here $v^t$, $W_{eh}$, $W_{sh}$ and $b_{att}$ are model parameters to be learned, and $h_i$ is the concatenation  of forward and backward hidden states of the encoder. 
We use this attention $a_i^t$ to generate the \emph{context vector} $c_t^*$ as a weighted sum of encoder hidden states: $c_t^* = \sum_i a_i^th_i$. We further use the $c_t^*$ vector to obtain a probability distribution over the words in the vocabulary as: $P = sofmax(W_v[s_t,c_t^*]+b_v)$,
where $W_v$ and $b_v$ are model parameters. Thus during decoding, the probability of a word is $P(qword)$. 
During the training process for each timestamp, the loss is calculated as $L_t = -\log P(qword_t)$. The loss associated with the  generated question is:
\begin{equation}
Loss = \frac{1}{T}\sum_{t=0}^{T}L_t = -\frac{1}{T}\sum_{t=0}^{T} \log P(qword_t)
\end{equation}

\subsubsection{The Copy and Coverage Mechanisms:}
\label{cnc}
The copy mechanism facilitates the copying of important entities and words from the source sentence to the question. 
We calculate $p_{cg} \in [0,1]$ as the decision of a binary classifier that determines whether to generate (sample) a word from the vocabulary or to copy the word directly from the input text, based on attention distribution $a_i^t$:
\begin{equation}
p_{cg} = sigmoid(W_{eh}^Tc_t^*+W_{sh}^Ts_t+W_{x}x_t+b_{cg})
\end{equation}
Here $W_{eh}$, $W_{sh}$, $W_{x}$ and $b_{cg}$ are trainable model parameters. 
The final probability of decoding a word is specified by the mixture model:
\begin{equation}
\label{pd}
p^*(qword) = p_{cg}\sum_{i:w_{i}=qword} a_i^t + (1-p_{cg})p(qword)
\end{equation}
Where $p^*(qword)$ is the final distribution over the union of the vocabulary and the input sentence.

As discussed earlier, Equation (\ref{pd}) addresses the rare words issue, since a word not in vocabulary will have probability $p(qword)=0$. Therefore, in such cases, our model will replace the {$<$unk$>$} token for out-of-vocabulary words with a word in the input sentence having the highest attention obtained using attention distribution $a_i^t$.

To discourage meaningless multiple repetitions of words in the question (as illustrated in row 3 of Table~\ref{tab:sample}), we maintain a word coverage vector ($wcv$) for the words already predicted as the sum of all the attention distributions ranging over timesteps $0$ until $t-1$. Specifically, at time step $t$, $wcv = \sum^{t-1}_{t'=0} a^{t'}$. 

No word is generated before timestep $0$, and hence \textit{wcv} will be a zero vector then. 
After storing the word coverage vector until $t-1$, while attending to the next word, we will need to inform our attention mechanism  about words covered until then. Hence, equation (\ref{att}) is now modified to be:
\begin{equation}
e^t_i = v^ttanh(W_{wcv}wcv_i^t+W_{eh}h_i+W_{sh}s_t+b_{att})
\end{equation}
Here $W_{wcv}$ are trainable parameters that inform the attention mechanism about  words that have been previously covered while choosing to attend over the next word. 
Following the incorporation of the copy and coverage mechanism in our generator, the generator's final loss function will be:
\begin{equation}
\label{finalloss}
Loss_{copy+cov}= \frac{1}{T}\sum_{t=0}^{T}\log P^*(w_t)-\lambda_cL_{cov}
\end{equation}
where $\lambda_c$ is the coverage hyperparameter and the coverage loss $L_{cov}$ is defined as:
\begin{equation}
L_{cov}=\sum_i min(a_i^t,wcv_i^t)
\end{equation}
We note that this cross-entropy based loss function still does not include task-specific metrics such as BLEU that were motivated earlier. 
We employ an evaluator to refine the model pre-trained on this loss function to directly optimize the task specific reward. We also empirically show that the refinement of maximum likelihood models using task-specific rewards such as BLEU improves results considerably. In the next subsection we describe our evaluator.

\subsection{Evaluator}
\label{eval}
The evaluator fine-tunes the parameters of the generator network by optimizing task-specific reward functions through policy gradient. It takes as input the predicted sequence and the gold sequence, evaluates a policy, and returns a reward (a score between $0$ and $1$) that reflects the quality of the question generated. For question generation, the choice of reward functions include task-specific metrics BLEU, GLEU and ROUGE-L~\cite{du2017learning,kumarpakdd2018}, as well as the decomposable attention~\cite{parikh2016decomposable} described below. More importantly, we present two new reward functions that are specifically designed for question generation, QSS and ANSS, for the conformity of questions and answers respectively. 

Combining Equation (\ref{finalloss}) with a reward function $R$ (BLEU, GLEU, ROUGE, DAS, QSS and ANSS), we obtain the overall loss function using the expected reward objective as follows:
\begin{align}
\begin{split}
    L_{overall} = & \alpha * Loss_{copy+cov} \\
    &\quad + \beta * \sum\limits_{i=0}^N \sum_{y \in \mathcal{Y}}P_{\theta}(y|X^{(i)})R(y,y^{*(i)})
\end{split}
\label{overall_loss} 
\end{align}
where $R(y,y^{*(i)})$ denotes per sentence score (reward), $\mathcal{Y}$ is a set of sequences sampled from the final distribution, and $\alpha$ and $\beta$ are tunable hyperparamters.

\subsubsection{Decomposable attention based evaluator}
\label{das}
The use of a lexical similarity based reward function such as BLEU or ROUGE does not provide the flexibility to handle multiple possible versions of the ground truth. For example, the questions \textit{``who is the widow of ray croc?''} and \textit{``ray croc was married to whom?''} have almost the same meaning, but due to word order mismatch with the gold question, at most one of them can be rewarded using the BLEU score at the cost of the other(s). Empirically, we find this restriction leading to models that often synthesize questions with poor quality.
 We therefore, design a novel reward function, a decomposable attention~\cite{parikh2016decomposable} based similarity scorer (DAS).
Denoting by $\hat{q}$ a generated question and by $q$ the ground-truth question, we compute a cross attention based similarity using the following steps: 
\paragraph{Cross Attention:} The generated question $\hat{q}$ and the ground-truth question $q$ are inter-attended as:
\begin{equation}
\begin{split}
\hat{q}^*_i &= \sum\limits_{j=0}^{L_{q}} a_{ji}e(q_j),\ a_{ji} = \frac{\exp (e(\hat{q}_i)^Te(q_j))}{\sum_{k=0}^{L_{\hat{q}}}\exp(e(\hat{q}_i)^Te(q_k))}, \\
q^*_j &= \sum\limits_{i=0}^{L_{\hat{q}}} b_{ji}e(\hat{q}_i),\ b_{ji} = \frac{\exp(e(\hat{q}_i)^Te(q_j))}{\sum_{k=0}^{L_{q}}\exp(e(\hat{q}_k)^Te(q_j))}
\end{split}
\end{equation}
where $e(.)$ is the word embedding of dimension size $d$, $\hat{q}^*$ is the cross attention vector for a generated question $\hat{q}$, and $q^*$ is the cross attention vector for a   question $q$ in the ground truth.
\paragraph{Comparison:}  Each n-gram $\hat{q}_i$ in the generated question  (through its embedding $e(\hat{q}_i)$) is compared with its associated cross-attention vector $\hat{q}^*$ using a feed forward neural network $N_1$. Similarly,  each n-gram  $q_j$ in the ground-truth question (through its embedding $e(q_j)$) is compared with its associated attention vector $q^*$ using another network $N_2$ having the same architecture as $N_1$. The motivation for this comparison is that we would like to determine the soft alignment between  n-grams in the generated question and the gold question. As an illustration, while comparing the gold question ``\emph{why do rockets look white?}'' with a generated question ``\emph{why are rockets and boosters painted white?}'', we find  that an n-gram ``\emph{rockets and boosters}'' is softly aligned to ``\emph{rockets}'' while ``\emph{look}'' is softly aligned to ``\emph{painted}''.
\begin{equation}
\bm{\hat{q}_{1,i}} = N_1([e(\hat{q}_i),\hat{q}^*]),\ \bm{q_{2,j}} = N_2([e(q_j),q^*])\label{fm:das}
\end{equation}
where $\bm{\hat{q}_{1,i}}$ and $ \bm{q_{2,j}}$  are vectors containing comparison scores of aligned phrases in generated question and gold question respectively and $N_1$ and $N_2$ are the feed forward neural nets.
\paragraph{Matching Score:} 
 The vectors \bm{$\hat{q}_{1,i}$} and \bm{$q_{2,j}$} are aggregated over each word or phrase in the predicted question and  gold question respectively before feeding them to a linear function ($L$):
\begin{equation}
DAS = L(\sum_{i=1}^{L_q}\bm{\hat{q}_{1,i}},\sum_{j=1}^{L_{\hat{q}}}\bm{q_{2,j}})
\label{eqn:match}
\end{equation}
This matching score between the predicted question and the gold question is the reward returned by the decomposable attention based evaluator.
\subsubsection{QG quality specific reward functions}
\label{qgquality}
We introduce two new reward functions that specifically designed to evaluate the conformity of the generated question (QSS) and answer (ANSS) against the ground truth.

\textbf{Question sentence overlap score (QSS): } This reward function is specific to QG. We compute the sentence overlap score as the number of common n-grams between predicted question and the source sentence. This reward ensures that generated question is relevant to the given sentence. 
Thus, if $precision_n(s,q)$ computes the $n-$gram precision match between sentence and question,\\
\begin{equation}
    QSS=(\prod_{i=1}^n precision_i(sentence, question))^{\frac{1}{n}}
\end{equation}

\textbf{Predicted and encoded answer overlap score (ANSS):} In order to ensure that the generated question is about the pivotal answer/ground truth answer we calculate answer overlap score. Answer overlap score is the number of common n-grams between the encoded answer and the answer predicted (ans$_{qa}$) for the generated question using the best performing question answering model over SQuAD\footnote{\url{https://github.com/huggingface/pytorch-pretrained-BERT}}

\begin{align}
    ANSS=(\prod_{i=1}^n precision_i(\text{ans$_{qa}$}, pivotal\_answer))^{\frac{1}{n}}
\end{align}


\section{Experimental Setup}
\label{exp}
In this section, we present our evaluation framework on the publicly available SQuAD~\cite{rajpurkar-EtAl:2016:EMNLP2016} dataset. We first explain various reward functions employed in our experiments. We then describe our baseline and the evaluation methods. 

\paragraph{Reward Functions:}
We experimented with the five reward functions discussed in Section \ref{eval}: (1) BLEU, (2) GLEU, (3) ROUGE-L, (4) DAS, and (5) the QG-specific reward QSS+ANSS. 
In our experiments we considered BLEU for up to 4-grams. For the GLEU score, we recorded all sub-sequences of up to $4$-grams. 

\paragraph{Baselines and Evaluation Methods:}
We reimplemented two state-of-the-art question generation models as baselines for comparison: L2A~\cite{du2017learning} and AutoQG~\cite{kumarpakdd2018}. A direct (and fair) comparison with another recent technique, NQG$_{LC}$~\cite{song2018leveraging}, is not feasible, as unlike us, NQG$_{LC}$ requires ground-truth answers, whereas both AutoQG and our model predict pivotal answers. L2A does not consider answers. Moreover, their context (input is sometimes more than one sentence) is different also the train/test split is different from ours. Hence, we only report the original numbers reported in their paper. We also did not perform human evaluation on NQG$_{LC}$ as their source code has not been made available for reimplementation.

We also use an existing implementation of a recent RL-based abstractive summarization technique \cite{paulus2018} to train baseline models \sumbleu\ (with BLEU as reward function) and \sumrouge\ (with ROUGE as reward function). This comparison studies the effectiveness of state-of-the-art abstractive summarization techniques applied to question generation as-is, as the two are conceptually similar tasks.

We report automatic and human evaluation results on eight variants of our model, each of which is equipped with the copy and coverage mechanism, the pointer network, as well as one of the four reward functions: BLEU, GLEU, ROUGE-L, DAS or one of the four rewards in combination with QG quality specific rewards (QSS+ANSS). Hence, our models are named \rlbleu, etc. 

For automatic evaluation, we employ BLEU, ROUGE-L and METEOR, which are standard evaluation measures used to evaluate sequence prediction tasks. We use the evaluation scripts released by \cite{chen2015microsoft} that was originally used to evaluate the image captioning task.

We also performed human evaluation to further analyze the quality of questions generated for their syntactic correctness, semantic correctness and relevance. Syntactic correctness measures the grammatical correctness of a generated question, semantic correctness measures meaningfulness and naturalness of the question, and relevance measures how relevant the question is to the text. We perform human evaluation for each model on a randomly selected subset of 100 sentences. Each of the three judges is presented the 100 sentence-question pairs for each model and asked for a binary response on each quality parameter. The responses from all the judges for each parameter is then averaged for each model. 

\begin{table*}[htb]
\begin{center}
\scalebox{0.8}{
  \begin{tabular}{| l | c | c |c | c | c | c | }
    \hline
     Model& BLEU-1 & BLEU-2 & BLEU-3 & BLEU-4 &METEOR &ROUGE-L \\ \hline
     L2A~\cite{du2017learning} &43.21 (43.09) &24.77 (25.96) &15.93 (17.50) &10.60 (12.28) &16.39 (16.62) &38.98 (39.75)\\
     AutoQG~\cite{kumarpakdd2018} &44.68 (46.32) &26.96 (28.81) &18.18 (19.67) &12.68 (13.85) &17.86 (18.51) &40.59 (41.75) \\
     NQG$_{LC}$~\cite{song2018leveraging} & - & - & - & - (13.98) & - (18.77) & - (42.72)\\
     \sumbleu~\cite{paulus2018} &11.20- &3.50- &1.21- &0.45- &6.68- & 15.25-\\
     \sumrouge~\cite{paulus2018} &11.94- &3.95- &1.65- &0.082- &6.61- &16.17- \\
     \hline
     \rlbleu &46.84 &29.38 &20.33 &14.47 &19.08 &41.07\\
      \rowcolor{Gray}
      \rlbleuqa &46.59 &29.68 &20.79 &15.04 &19.32 &41.73\\
     \rldas &44.64  &28.25 &19.63 &14.07 &18.12 &42.07\\
      \rowcolor{Gray}
     \rldasqa &46.07  & 29.78 & 21.43 & 16.22 & 19.44& 42.84\\
     \rlgleu & 45.20&29.22&20.79 & 15.26&18.98 &43.47\\
      \rowcolor{Gray}
    \rlgleuqa &47.04 &30.03 &21.15 & 15.92 & 19.05 & 43.55 \\  
     \rlrouge &47.01 &30.67 &21.95 &16.17 &19.85 &43.90\\
      \rowcolor{Gray}
     \rlrougeqa & \textbf{48.13} & \textbf{31.15} & \textbf{22.01} & \textbf{16.48} & \textbf{20.21}& \textbf{44.11}\\
     \hline
  \end{tabular}}
\end{center}

\caption{Experimental results on the test set on automatic evaluation metrics. Best results for each metric (column) are \textbf{bolded}. The numbers in parentheses for L2A, AutoQG and NQG$_{LC}$ are those from the best models reported in their respective original papers. The slight difference of up to 1.7\% from our reproduced numbers can be attributed to reimplementation and different versions of various libraries used. Models with new QG-specific reward functions (QSS+ANSS) are highlighted in gray for easy comparison.}
\label{results}

\end{table*}

\subsection{Ablation Analysis}
\begin{table*}[ht]
\begin{center}
\small
\scalebox{0.8}{
\begin{tabular}{| l | c | r |c | c | c | c | }
    \hline
     \makecell{\textbf{Model} \\ (\textbf{\rlrouge})}& \makecell{\textbf{$\Delta$ BLEU-1} \\ \textbf{(47.01)}} & \makecell{\textbf{$\Delta$ BLEU-2}\\ \textbf{(30.67)}} & \makecell{\textbf{$\Delta$ BLEU-3} \\ \textbf{(21.95)} }& \makecell{\textbf{$\Delta$ BLEU-4}\\ \textbf{(16.17)}} &\makecell{\textbf{$\Delta$ METEOR} \\ \textbf{(19.85)}} &\makecell{\textbf{$\Delta$ ROUGE-L} \\ \textbf{(43.90)} }\\ \hline
     W/o copy &2.09 (4.7\%) & 2.13 (6.9\%) &2.94 (13.4\%) & 2.23 (13.8\%) &2.21 (11.1\%) & 2.58 (5.9\%) \\ \hline
     W/o coverage & 0.31 (0.7\%) & 0.57 (1.9\%) & 0.94 (4.2\%) & 0.28 (1.7\%) & 0.84 (4.2\%) & 1.01 (2.3\%) \\ \hline
\end{tabular}}
\end{center}

\caption{Ablation analysis results after removing (a) copy mechanism and (b) coverage mechanism from the system (\rlrouge). Both absolute performance drop and percentage of drop (in parentheses) are reported.}
\label{ablation}

\end{table*}

We conducted an ablation analysis to study the effect of removing the copy and coverage mechanisms. Table~\ref{ablation} summarizes the drop in performance for \rlrouge. Without the copy mechanism, there is a drop overall in every evaluation measure, with BLEU-4 registering the largest drop of 13.8\% as against 13.4\%, 6.9\% and 4.7\% in BLEU-3, BLEU-2 and BLEU-1 respectively. On the other hand, without the coverage mechanism, we see a consistent but sufficiently lower drop (1-2\%) in each evaluation measure for \rlrouge.


\section{Results and Discussion}
\label{rnd}
We show and compare results on automatic evaluation in Table~\ref{results}. Note the numbers in parentheses for L2A~\cite{du2017learning}, AutoQG~\cite{kumarpakdd2018}, and NQG$_{LC}$~\cite{song2018leveraging} are those reported in their original papers. The slight difference of up to 1.7\% in the original and reproduced numbers can be attributed to reimplementation and different versions of various libraries used. As can be seen, all our eight models outperform L2A and AutoQG on all evaluation metrics. Two of our models, \rlgleu\ and \rlrouge, also outperform NQG$_{LC}$. Hence, using evaluation metrics as the reward function during reinforcement based learning improves performance for all metrics. We also observe that \rlrougeqa, the model reinforced with ROUGE-L (that measures the longest common sequence between the ground-truth question and the generated question) as the reward function in combination with QG quality specific rewards(QSS+ANSS), is the best performing model on all metrics, outperforming existing baselines considerably. For example, it improves over AutoQG on BLEU-4 by 29.98\%, on METEOR by 13.15\%, and on ROUGE-L by 8.67\%.

In Table \ref{heresults} we present human evaluation results for the models evaluated on three quality parameters (a) syntactic correctness, (b) semantic correctness, and (c) relevance. 

Consistent with automatic evaluation results shown in Table~\ref{results}, seven of our eight models outperform the two baselines, with \rldasqa{} being the best model on syntactic correctness and semantic correctness quality metrics, outperforming all the other models by a large margin. However, model \rlbleuqa{} generates highly relevant questions and is the best model on relevance metrics. 

It is noteworthy that for each of our models (e.g.\ \rlbleu), adding QG-specific rewards (e.g.\ \rlbleuqa) significantly improves question quality in human evaluation, even though there is less noticeable improvements in automatic evaluation. This clearly demonstrates the effectivess of our new QG-specific reward functions.

We measure inter-rater agreement using Randolph's free-marginal multirater kappa~\cite{randolph2005free}. This helps in analyzing level of consistency among observational responses provided by multiple judges. It can be observed that our quality metrics for all our models are rated as \emph{moderate agreement}~\cite{viera2005understanding}.

\subsection{Analyzing Choice of Reward Function}
BLEU\cite{papineni2002bleu} measures precision and ROUGE\cite{lin2004rouge} measures recall, we believe that cross-entropy loss was already accounting for precision to some extent and using it in conjunction with ROUGE (which improves recall) therefore gives best performance.

\begin{table}[htb]
\begin{center}
\small
\scalebox{0.8}{
  \begin{tabular}{| l | c | c | c | c |c | c | }
    \hline
     \multirow{2}{*}{Model}& \multicolumn{2}{c|}{Syntax} & \multicolumn{2}{c|}{Semantics} & \multicolumn{2}{c|}{Relevance} \\ \cline{2-7}
            & Score & Kappa & Score & Kappa & Score & Kappa \\
    \hline
     L2A &39.2 &0.49 & 39 &0.49 &29 & 0.40  \\
     AutoQG &51.5 &0.49 &48 &0.78 &48 & 0.50  \\
     \hline
     \rlbleu &47.5 &0.52 &49 &0.45 &41.5 &0.44 \\
     \rowcolor{Gray}
     \rlbleuqa & 82 &0.63 &75.3 &0.68 & \textbf{78.33}& 0.46\\
     \rldas &68 &0.40 &63  & 0.33 &41 & 0.40 \\
     \rowcolor{Gray}
     \rldasqa & \textbf{84} &0.57 & \textbf{81.3} &0.60 &74 & 0.47 \\
     \rlgleu & 60.5 &0.50 &62 & 0.52 &44 & 0.41 \\
     \rowcolor{Gray}
     \rlgleuqa & 78.3 &0.68  & 74.6 &0.71  & 72& 0.40 \\
     \rlrouge & 69.5& 0.56 & 68 & 0.58 & 53 & 0.43 \\
     \rowcolor{Gray}
     \rlrougeqa & 79.3 &0.52 & 72 &0.41 & 67 & 0.41 \\
      \hline
  \end{tabular}}
\end{center}
\caption{Human evaluation results (column ``Score'') as well as inter-rater agreement (column ``Kappa'') for each model on the test set. The scores are between 0-100, 0 being the worst and 100 being the best. Best results for each metric (column) are \textbf{bolded}. The three evaluation criteria are: (1) syntactically correct ({Syntax}), (2) semantically correct ({Semantics}), and (3) relevant to the text ({Relevance}). Models with new QG-specific reward functions (QSS+ANSS) are highlighted in gray for easy comparison.}
\label{heresults}

\end{table}

DAS calculates semantic similarity between generated question and the gound-truth question. 
As discussed in section \ref{das} DAS will give high reward even though the generated question has low BLEU score. Thus, the performance of the model on automatic evaluation metrics does not improve with  DAS as the reward function, though the quality of questions certainly improves. Further, ROUGE in conjunction with the cross entropy loss improves on recall as well as precision whereas every other combination overly focuses only on precision.

Error analysis of our best model reveals that most errors can be attributed to intra-sentence dependencies such as co-references, concept dependencies {\em etc.} In a camera ready version of the paper, we will share link to a detailed report containing extensive experiments that include ablation tests. Also link to the source code will be provided then.


\section{Related Work}
\label{relwork}
Neural network-based methods represent the state-of-the-art in automatic question generation (QG) from text. Motivated by neural machine translation, Du et al \shortcite{du2017learning} proposed a sequence-to-sequence (Seq2Seq) architecture for QG. In our previous work, we \shortcite{kumarpakdd2018} proposed to augment each word with linguistic features and encode the most relevant \emph{pivotal answer} to the text while generating questions. Similarly, Song et al \shortcite{song2018leveraging} encode ground-truth answers (given in the training data), use the copy mechanism and additionally employ context matching to capture interactions between the answer and its context within the passage. They encode ground truth answer for generating questions which might not be available for test set in contrast we train a Pointer Network based model to predict the pivotal answer to generate question about. In our work \cite{Kumar2019Difficulty} we proposed a transformer based architecture to automatically generate complex multi-hop questions from knowledge graphs. In \cite{Kumar2019CrossLingualTF} we proposed a cross lingual training method for automatically generating questions from text in low resource languages.

\par Very recently deep reinforcement learning has been successfully applied to natural language generation tasks such as abstractive summarization~\cite{paulus2018,N18-1150} and dialogue generation~\cite{D16-1127}. In summarization, one generates and paraphrases sentences that capture salient points of the text. On the other hand, generating questions additionally involves determining question type such as what, when, etc., being selective on which keywords to copy from the input into the question, leaving remaining keywords for the answer. This also requires the development of a specific probabilistic generative model. \cite{yao2018teaching} proposed generative adversarial network (GAN) framework with modified discriminator to predict question type. Recently Fan et al \shortcite{fan2018reinforcement} proposed a bi-discriminator framework for visual question generation. They formulate the task of visual question generation as a language generation task with some linguistic and content specific attributes.


\section{Conclusion}
\label{conc}
We presented a novel, holistic treatment of question generation (QG) using a generator-evaluator framework. Our  generator provisions for explicitly factoring in question syntax and semantics, identifies pivotal answers, recognizes contextually important words and avoids meaningless repetitions. Our evaluator allows us to directly optimize for conformity towards the structure of ground-truth question(s). We propose two novel reward functions account for conformity with respect to ground-truth questions and predicted answers respectively. In conjunction, the evaluator makes use of task-specific scores, including BLEU, GLEU, ROUGE-L, and decomposable attention (DAS) that are naturally suited to QG and other seq2seq problems. Experimental results on automatic evaluation and human evaluation on the standard benchmark dataset show that our framework, especially with the incorporation of the new reward functions, considerably outperforms state-of-the-art systems.

\bibliography{emnlp-ijcnlp-2019}
\bibliographystyle{acl_natbib}
\end{document}